\documentclass{article} 
\usepackage[preprint]{colm2026_conference}

\usepackage{microtype}
\usepackage{hyperref}
\usepackage{url}
\usepackage{booktabs}
\usepackage{amsmath}
\usepackage{tabularx}
\usepackage{graphicx}
\usepackage{xcolor}

\usepackage{lineno}

\definecolor{darkblue}{rgb}{0, 0, 0.5}
\hypersetup{colorlinks=true, citecolor=darkblue, linkcolor=darkblue, urlcolor=darkblue}

\title{Read More, Think More: \\Revisiting Observation Reduction for Web Agents}

\author{Masafumi Enomoto, Ryoma Obara, Haochen Zhang \& Masafumi Oyamada \\
NEC Corporation\\
\texttt{\{masafumi-enomoto,ryoma-obara,haochen-zhang,oyamada\}@nec.com}
}

\begin{document}

\ifcolmsubmission
\linenumbers
\fi

\maketitle

\begin{abstract}
Web agents based on large language models (LLMs) rely on observations of web pages---commonly represented as HTML---as the basis for identifying available actions and planning subsequent steps.
Prior work has treated the verbosity of HTML as an obstacle to performance and adopted observation reduction as a standard practice. We revisit this trend and demonstrate that the optimal observation representation depends on model capability and thinking token budget:
(1) compact observations (accessibility trees) are preferable for lower-capability models, while detailed observations (HTML) are advantageous for higher-capability models; moreover, increasing thinking tokens further amplifies the benefit of HTML.
(2) Our error analysis suggests that higher-capability models exploit layout information in HTML for better action grounding, while lower-capability models suffer from increased hallucination under longer inputs.
We also find that incorporating observation history improves performance across most models and settings, and a diff-based representation offers a token-efficient alternative.
Based on these findings, we suggest practical guidelines: adaptively select observation representations based on model capability and thinking token budget, and incorporate observation history using diff-based representations.
\end{abstract}

\section{Introduction}
\label{sec:introduction}
LLM-based web agents have attracted growing research interest 
because they can automate repetitive web tasks such as form filling 
and ordering items~\citep{related_work/webagent_survey}.
Web agents are generally implemented as a loop of action execution 
and observation~\citep{react}: at each step, the agent 
executes an action (e.g., a click) and receives the current web page 
as an observation.
This loop continues until the agent judges the task complete.
The information contained in the web page observation is critical 
for identifying available next actions and tracking task progress.

\begin{figure}[t]
\centering
\includegraphics[width=0.90\columnwidth]{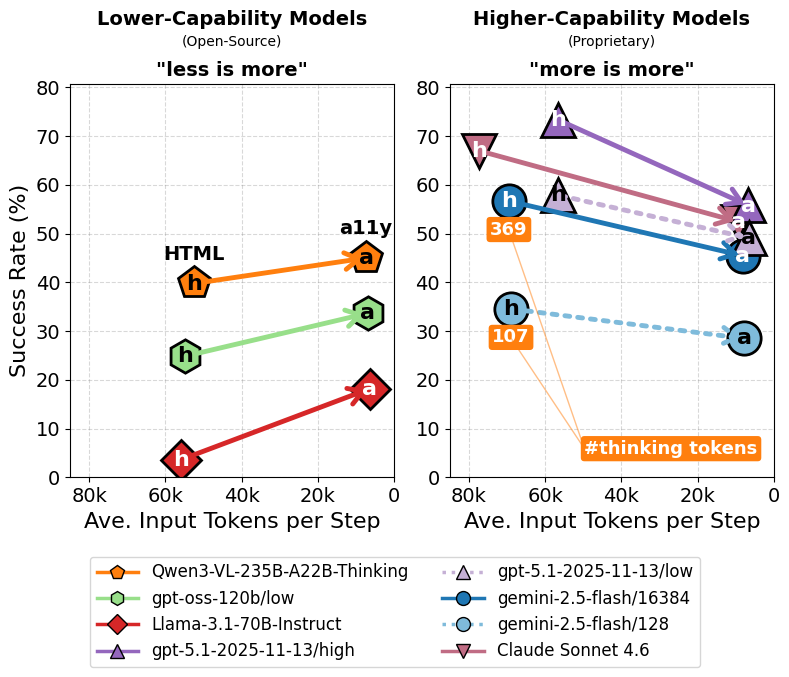}
\caption{Relationship between observation representation and task success rate (WorkArena L1). The x-axis shows the average number of input tokens per step; the y-axis shows the task success rate. ``h'' denotes HTML and ``a'' denotes the accessibility tree (a11y). Arrows indicate the change from HTML to a11y within the same model. \textbf{Left:} For lower-capability (open-source) models, reducing the observation (h $\rightarrow$ a) improves the success rate. \textbf{Right:} For higher-capability (proprietary) models, the success rate decreases instead.}
\label{fig:killer}
\end{figure}

A consistent trend in existing web agent research is to \emph{reduce} the amount of observation information.
The primary motivation is that HTML representations of web pages tend to be extremely long---for example, on WorkArena (a web agent benchmark), HTML pages range from a minimum of 40K to a maximum of 500K tokens~\citep{benchmark/workarena}.
Early LLMs could not directly take such long inputs due to context window limits~\citep{related_work/mind2web}.
LLMs with million-token context windows, such as Gemini,\footnote{\url{https://gemini.google.com/}} have emerged and can accept raw HTML directly.
However, observation reduction remains dominant because task-irrelevant information impedes reasoning~\citep{related_work/agent_occam}.
Various reduction strategies have been proposed, including extracting only task-relevant elements~\citep{related_work/mind2web,related_work/html-t5,related_work/weblinx}, converting pages to compact representations~\citep{related_work/agent_occam,related_work/lcow}, and adopting the accessibility tree (a11y) as a more compact alternative to HTML~\citep{related_work/agent_s,related_work/oscar}.
When historical observations from past steps are also used, reduction is similarly applied~\citep{related_work/agent_occam,related_work/autowebglm}.

In this work, we revisit the trend of observation reduction.
The conventional rationale for reduction was rooted in the limited long-context capability of earlier LLMs.
Recent LLMs, however, have demonstrated substantially improved ability to process long inputs~\citep{related_work/LCLM_reasoning/beyond_isolated}, and extended chain-of-thought reasoning at inference time further improves their capability~\citep{related_work/LCLM_reasoning/cot_matters}.
Motivated by these advances, we investigate the following research question using state-of-the-art LLMs:
\emph{How does the effect of observation reduction vary with model capability and the amount of thinking tokens at inference time?}

Our experiments on WorkArena L1~\citep{benchmark/workarena} reveal that \textbf{observation reduction is not universally beneficial---the optimal observation representation depends on model capability and thinking token budget}:
compact observations (a11y) are preferable for lower-capability models, while detailed observations (HTML) are advantageous for higher-capability models (Figure~\ref{fig:killer}, \S\ref{sec:experiment/modality})~\footnote{In our experimental setup, proprietary models serve as a proxy for higher-capability models and open-source models for lower-capability models.}
; moreover, increasing thinking tokens further amplifies the benefit of HTML.
Action analysis suggests that higher-capability models exploit layout information in HTML for better action grounding, while lower-capability models suffer from increased hallucination under longer inputs (\S\ref{sec:experiment/grounding_error}).
Incorporating observation history improves performance across most models and settings, and a diff-based representation offers a token-efficient alternative (\S\ref{sec:experiment/history}).
Based on these findings, we suggest practical guidelines for web agent system design: we recommend \textbf{adaptively selecting observation representations based on model capability and thinking token budget}, and encourage \textbf{incorporating observation history with diff-based representations as a token-efficient alternative}.

\section{Problem Setting}
\label{sec:problem_setting}
Following~\citet{related_work/agent_occam}, we formalize the web agent task as a partially observable Markov decision process (POMDP): $\langle S, A, O, P, R, \Omega \rangle$, where $S$ is the state space, $A$ is the action space, $O$ is the observation space, $P: S \times A \rightarrow S$ is the state transition function, $R: S \times A \times S \rightarrow \mathbb{R}$ is the reward function, and $\Omega: S \rightarrow O$ is the observation function.

An LLM-based agent operates as follows.
At each step $t$, the agent selects an action $a_t \in A$ according to the policy $\pi_{\text{LLM}}(a_t \mid I, o_t, \{(o_i, a_i)\}_{i=0}^{t-1})$, where $I$ is the system prompt containing the task instruction, $o_t$ is the current observation, and $\{(o_i, a_i)\}_{i=0}^{t-1}$ is the history of past observation--action pairs.
Upon executing action $a_t$, the environment transitions to the next state $s_{t+1} = P(s_t, a_t)$ according to the transition function, and the agent receives a new observation $o_{t+1} = \Omega(s_{t+1})$ via the observation function.
The observation $o_{t+1}$ is provided as a textual representation of the web page (HTML or accessibility tree) or as a rendered image (screenshot).
This process repeats until the agent executes a termination action or the maximum number of steps $T_{\max}$ is reached.
Task success is evaluated by the reward $r = R(s_T, a_T, s_{T+1})$ upon episode termination, where $T$ denotes the terminal step.

In this work, we vary the following two aspects of how observations are fed into the policy $\pi_{\text{LLM}}$:
\begin{enumerate}
  \item \textbf{Observation representation}: The current observation $o_t$ is represented as HTML, an accessibility tree (a11y), a screenshot, or a combination thereof.
  \item \textbf{Observation history length}: The number of past observations $\{o_i\}_{i<t}$ included as history.
\end{enumerate}
We evaluate these observation design choices in combination with different language models and varying amounts of thinking tokens.

\section{Experiments}
\label{sec:experiment}
\subsection{Experimental Setup}
\label{sec:experiment/setting}
\noindent
\textbf{Benchmark.}
We evaluate on \textbf{WorkArena}~\citep{benchmark/workarena, benchmark/workarena++}, a web agent benchmark built on the ServiceNow platform\footnote{\url{https://www.servicenow.com/}} that assesses the ability to perform everyday workplace tasks such as ordering items and filling out forms.
We use WorkArena L1, which consists of 330 tasks in total: 33 task types $\times$ 10 seeds.

\noindent
\textbf{Language Models.}
To assess the effect of observation design, we consider a diverse set of model configurations along the following axes:
(1) \textit{Model capability}: we compare higher-capability frontier proprietary models against lower-capability open-source models, and compare models of different sizes within the same family (e.g., \texttt{gpt-oss-120b} vs.\ \texttt{gpt-oss-20b}).
(2) \textit{Amount of thinking tokens}: we vary the parameter that controls the amount of thinking tokens for each reasoning model where applicable.
For proprietary models, we use \texttt{claude-sonnet-4-6}~\footnote{\url{https://www.anthropic.com/news/claude-sonnet-4-6}}, \texttt{gpt-5.1-2025-11-13}~\footnote{\url{https://openai.com/index/gpt-5-1/}}, \texttt{o3-mini-2025-01-31}~\footnote{\url{https://openai.com/index/openai-o3-mini/}}, and \texttt{gemini-2.5-flash}~\citep{models/gemini_25}.
For open-source models, we use \texttt{gpt-oss-120b/20b}~\citep{models/gpt-oss}, the \texttt{Qwen3-VL} family (\texttt{235B-A22B-Thinking}, \texttt{30B-A3B-Thinking}, \texttt{30B-A3B-Instruct})~\citep{models/qwen3}, and the \texttt{Llama-3.1} family (\texttt{70B-Instruct}, \texttt{8B-Instruct})~\citep{models/llama3}.
For \texttt{gpt-5.1}, \texttt{o3-mini}, and \texttt{gpt-oss}, thinking tokens are controlled via the \texttt{reasoning\_effort} parameter (low/high); for \texttt{gemini}, via the \texttt{thinking\_budget} parameter (128/16384).
The \texttt{Llama-3.1} family is included to evaluate performance of older-generation models that are not trained as reasoning models.

\noindent
\textbf{Action Grounding.}
Grounding refers to the process of mapping the action string produced by the agent to an actual browser operation.
We adopt id-based grounding: the agent specifies an action $a_t$ by referencing the id of a DOM element present in the current observation $o_t$.
Each action is expressed as a string combining an action type, a target element id, and optional parameters (e.g., \texttt{fill('12',~'hello')}).
Actions are executed via Playwright,\footnote{\url{https://playwright.dev/}} a browser automation framework.
We use AgentLab~\citep{browsergym, benchmark/workarena}, a framework for running and managing web agent experiments.
The maximum number of steps is set to $T_{\max} = 15$ following the framework default; tasks that exceed this limit are treated as failures.

\begin{table}[t]
\centering
\small
\tabcolsep 3pt
\caption{Task success rate (\%) by observation representation and model, evaluated on Workarena L1. Numbers in parentheses indicate the difference from a11y. N/A indicates that the model does not support image input. Values in parentheses after model names indicate \texttt{reasoning\_effort} (high/low) or \texttt{thinking\_budget}. Input token counts are provided as reference values for \texttt{gpt-5.1 (high)}, including image inputs.}
\label{tab:modality_workarena_l1}
\begin{tabular}{l|rrr}
\hline
 & a11y & html & a11y+scrs \\
\#input tokens & 6720 & 56653 & 7446 \\
\hline
\multicolumn{4}{l}{\textbf{Higher-Capability Models (Proprietary)}} \\
\hline
Claude Sonnet 4.6 & 52.4 & 67.0 (\textcolor{blue}{+14.6}) & 53.6 (\textcolor{blue}{+1.2}) \\
gpt-5.1-2025-11-13 (high) & 55.8 & 73.3 (\textcolor{blue}{+17.5}) & 59.7 (\textcolor{blue}{+3.9}) \\
gpt-5.1-2025-11-13 (low) & 49.1 & 57.9 (\textcolor{blue}{+8.8}) & 55.8 (\textcolor{blue}{+6.7}) \\
gemini-2.5-flash (budget=16384) & 45.5 & 56.7 (\textcolor{blue}{+11.2}) & 48.5 (\textcolor{blue}{+3.0}) \\
gemini-2.5-flash (budget=128) & 28.5 & 34.5 (\textcolor{blue}{+6.0}) & 32.7 (\textcolor{blue}{+4.2}) \\
o3-mini-2025-01-31 (high) & 39.7 & 32.1 (\textcolor{red}{-7.6}) & N/A \\
\hline
\multicolumn{4}{l}{\textbf{Lower-Capability Models (Open-Source)}} \\
\hline
gpt-oss-120b (high) & 46.7 & 38.8 (\textcolor{red}{-7.9}) & N/A \\
gpt-oss-120b (low) & 33.6 & 24.8 (\textcolor{red}{-8.8}) & N/A \\
gpt-oss-20b (high) & 46.4 & 27.6 (\textcolor{red}{-18.8}) & N/A \\
gpt-oss-20b (low) & 20.0 & 18.2 (\textcolor{red}{-1.8}) & N/A \\
Qwen3-VL-235B-A22B-Thinking & 45.0 & 39.8 (\textcolor{red}{-5.2}) & 44.8 (\textcolor{red}{-0.2}) \\
Qwen3-VL-30B-A3B-Thinking & 33.2 & 26.1 (\textcolor{red}{-7.1}) & 33.9 (\textcolor{blue}{+0.7}) \\
Qwen3-VL-30B-A3B-Instruct & 22.7 & 25.5 (\textcolor{blue}{+2.8}) & 24.8 (\textcolor{blue}{+2.1}) \\
Llama-3.1-70B-Instruct & 18.2 & 3.6 (\textcolor{red}{-14.6}) & N/A \\
Llama-3.1-8B-Instruct & 0.0 & 1.2 (\textcolor{blue}{+1.2}) & N/A \\
\hline
\end{tabular}
\end{table}

\begin{table}[t]
\centering
\small
\tabcolsep 3pt
\caption{Task success rate (\%) by observation format (Workarena L1). Comparing xml format vs indent format for both a11y and html representations. For each format group, the a11y column shows the baseline success rate, and the html column shows the difference from a11y in parentheses.}
\label{tab:format_comparison_workarena_l1}
\begin{tabular}{l|rr|rr}
\hline
 & \multicolumn{2}{c|}{xml} & \multicolumn{2}{c}{indent} \\
 & a11y & html & a11y & html \\
\hline
\multicolumn{5}{l}{\textbf{Higher-Capability Models (Proprietary)}} \\
\hline
gpt-5.1-2025-11-13 (high) & 60.0 & 73.3 (\textcolor{blue}{+13.3}) & 55.8 & 76.7 (\textcolor{blue}{+20.9}) \\
gemini-2.5-flash (budget=16384) & 51.5 & 56.7 (\textcolor{blue}{+5.2}) & 45.5 & 48.2 (\textcolor{blue}{+2.7}) \\
\hline
\multicolumn{5}{l}{\textbf{Lower-Capability Models (Open-Source)}} \\
\hline
gpt-oss-120b (high) & 50.0 & 38.8 (\textcolor{red}{-11.2}) & 46.7 & 26.7 (\textcolor{red}{-20.0}) \\
gpt-oss-20b (high) & 48.2 & 27.6 (\textcolor{red}{-20.6}) & 46.4 & 14.8 (\textcolor{red}{-31.5}) \\
\hline
\end{tabular}
\end{table}

\begin{table}[t]
\centering
\small
\tabcolsep 3pt
\caption{Impact of switching from a11y to HTML on task success rate by category (WorkArena L1). Values show the contribution to overall benchmark score (\%) when using HTML instead of a11y. Task categories: Form (record creation), Filter (list filtering), Sort (list sorting), Dashboard (chart reading), Navi (menu navigation), Knowledge (KB search), Catalog (service ordering).}
\label{tab:modality_category_diff}
\begin{tabular}{l|rrrrrrr}
\hline
 & Form & Filter & Sort & Dashboard & Navi & Knowledge & Catalog \\
\#tasks & 50 & 60 & 60 & 40 & 20 & 10 & 90 \\
\hline
\multicolumn{8}{l}{\textbf{Higher-Capability Models (Proprietary)}} \\
\hline
Claude Sonnet 4.6 & \textcolor{blue}{+1.2} & \textcolor{blue}{+11.8} & 0.0 & \textcolor{blue}{+1.5} & 0.0 & 0.0 & 0.0 \\
gpt-5.1-2025-11-13 (high) & \textcolor{red}{-1.8} & \textcolor{blue}{+8.8} & \textcolor{blue}{+6.7} & \textcolor{blue}{+3.0} & 0.0 & \textcolor{blue}{+0.3} & \textcolor{blue}{+0.6} \\
gpt-5.1-2025-11-13 (low) & \textcolor{red}{-4.5} & \textcolor{blue}{+7.0} & \textcolor{blue}{+3.9} & \textcolor{blue}{+2.7} & 0.0 & 0.0 & \textcolor{red}{-0.3} \\
gemini-2.5-flash (budget=16384) & \textcolor{red}{-1.8} & \textcolor{blue}{+6.1} & \textcolor{blue}{+5.5} & \textcolor{blue}{+3.3} & 0.0 & \textcolor{red}{-0.9} & \textcolor{red}{-0.9} \\
gemini-2.5-flash (budget=128) & \textcolor{red}{-2.4} & \textcolor{blue}{+4.2} & \textcolor{blue}{+3.0} & \textcolor{blue}{+1.2} & 0.0 & \textcolor{red}{-1.2} & \textcolor{blue}{+1.2} \\
o3-mini-2025-01-31 (high) & \textcolor{red}{-5.5} & \textcolor{blue}{+0.9} & \textcolor{blue}{+2.4} & \textcolor{blue}{+1.2} & 0.0 & \textcolor{red}{-1.2} & \textcolor{red}{-5.5} \\
\hline
\multicolumn{8}{l}{\textbf{Lower-Capability Models (Open-Source)}} \\
\hline
gpt-oss-120b (high) & \textcolor{red}{-3.9} & \textcolor{blue}{+1.8} & \textcolor{red}{-2.4} & \textcolor{blue}{+2.7} & 0.0 & 0.0 & \textcolor{red}{-6.1} \\
gpt-oss-120b (low) & \textcolor{red}{-3.0} & \textcolor{blue}{+0.6} & \textcolor{blue}{+0.6} & \textcolor{blue}{+0.9} & 0.0 & \textcolor{red}{-0.9} & \textcolor{red}{-7.0} \\
gpt-oss-20b (high) & \textcolor{red}{-5.5} & \textcolor{blue}{+0.9} & \textcolor{red}{-2.1} & \textcolor{blue}{+0.3} & 0.0 & \textcolor{red}{-1.2} & \textcolor{red}{-11.2} \\
gpt-oss-20b (low) & \textcolor{red}{-1.8} & 0.0 & 0.0 & \textcolor{blue}{+1.5} & 0.0 & \textcolor{red}{-0.6} & \textcolor{red}{-0.9} \\
Qwen3-VL-235B-A22B-Thinking & \textcolor{red}{-10.3} & \textcolor{blue}{+0.6} & \textcolor{blue}{+1.8} & \textcolor{red}{-5.8} & 0.0 & \textcolor{red}{-3.3} & \textcolor{red}{-33.6} \\
Qwen3-VL-30B-A3B-Instruct & \textcolor{red}{-1.8} & 0.0 & 0.0 & \textcolor{red}{-0.3} & 0.0 & \textcolor{red}{-0.6} & \textcolor{blue}{+5.5} \\
Qwen3-VL-30B-A3B-Thinking & \textcolor{red}{-9.4} & 0.0 & 0.0 & \textcolor{red}{-3.9} & 0.0 & \textcolor{red}{-2.7} & \textcolor{red}{-24.2} \\
Llama-3.1-70B-Instruct & \textcolor{red}{-2.4} & 0.0 & \textcolor{red}{-0.9} & \textcolor{red}{-2.7} & 0.0 & \textcolor{red}{-2.4} & \textcolor{red}{-6.1} \\
Llama-3.1-8B-Instruct & 0.0 & 0.0 & \textcolor{blue}{+0.9} & \textcolor{blue}{+0.3} & 0.0 & 0.0 & 0.0 \\
\hline
\end{tabular}
\end{table}

\subsection{Observation Representation and Task Success Rate}
\label{sec:experiment/modality}
We examine the effect of observation representation on task success rate.
The accessibility tree (a11y) is a compact representation that focuses primarily on interactive elements of the web page.
HTML, by contrast, preserves a more detailed document structure of the web page; in our experimental setting, it additionally includes CSS-based layout information.
Using a11y as the baseline, we evaluate two alternatives: (1) replacing a11y with HTML, and (2) augmenting a11y with screenshots (a11y+scrs).
Results are shown in Table~\ref{tab:modality_workarena_l1}.

\noindent
\textbf{Model Capability and Observation Representation.}
The results indicate that higher-capability models (proprietary in our setup) are better able to exploit more information-rich observations.
Higher-capability proprietary models (\texttt{claude-sonnet-4-6}, \texttt{gpt-5.1}, \texttt{gemini-2.5-flash}) perform better with HTML than with a11y, whereas lower-capability open-source models frequently degrade when HTML is used.
A similar trend holds for screenshots: augmenting a11y with screenshots yields larger gains for higher-capability models.
Within the same model family, smaller models tend to suffer greater performance degradation from HTML.
For example, with \texttt{reasoning\_effort=high}, \texttt{gpt-oss-20b} degrades more than \texttt{gpt-oss-120b}.
Even among proprietary models, older and more compact reasoning models such as \texttt{o3-mini} suffer from HTML, likely due to their limited long-context capability.

\noindent
\textbf{Thinking Tokens and Observation Representation.}
More thinking tokens lead to greater ability to exploit detailed observations such as HTML.
For instance, increasing \texttt{reasoning\_effort} from low to high for \texttt{gpt-5.1} yields a larger performance gain from HTML.
A similar trend is observed for \texttt{gemini-2.5-flash}, where increasing \texttt{thinking\_budget} from 128 to 16384 also leads to a larger gain from HTML.

\noindent
\textbf{XML vs. Indented Format.}
In Table~\ref{tab:modality_workarena_l1}, 
the a11y representation was provided in an indented format where hierarchy is expressed 
via space indentation, whereas HTML was provided in XML-tagged format.
This raises the possibility that the observed performance differences stem from 
the format choice rather than the information content of each representation.
For instance, if proprietary models are trained more heavily on XML-formatted data, 
their format preference may manifest in task performance.
To examine this possibility, we evaluated a11y and HTML in both XML and indented formats.
As shown in Table~\ref{tab:format_comparison_workarena_l1}, 
regardless of format, higher-capability models consistently benefit from HTML while lower-capability models consistently degrade.

\noindent
\textbf{Analysis by Task Category.}
Table~\ref{tab:modality_category_diff} shows the per-category change in task success rate when switching from a11y to HTML on WorkArena L1.
A consistent pattern emerges across models: HTML improves performance on Filter (list filtering), Sort (list sorting), and Dashboard (chart reading) tasks, while it degrades performance on Form (record creation), Knowledge (KB search), and Catalog (service ordering) tasks.
For higher-capability models, the gains on the former categories are large and the losses on the latter are small, resulting in an overall improvement with HTML.
For lower-capability models, the opposite pattern holds, leading to an overall degradation.
To better understand why higher-capability models achieve large gains on HTML-friendly categories while limiting losses on HTML-unfriendly ones, we analyze agent behavior at the step level in \S\ref{sec:experiment/grounding_error}.

\begin{figure}[t]
\centering
\includegraphics[width=0.90\columnwidth]{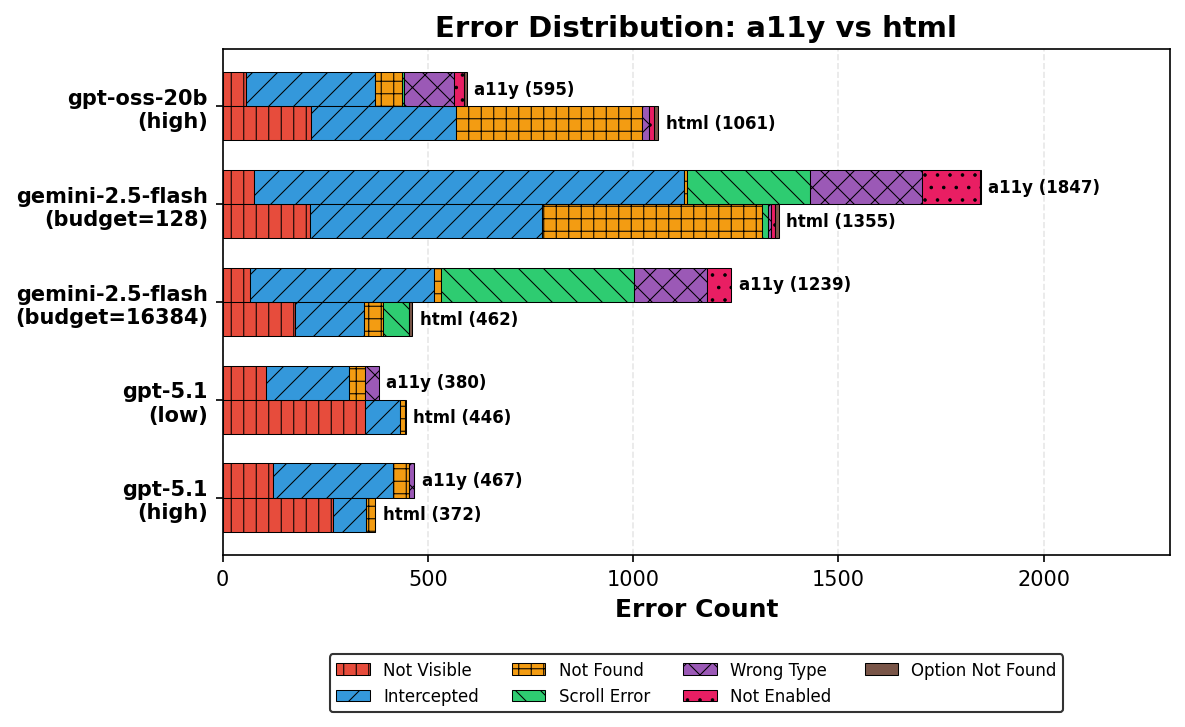}
\caption{Grounding error counts and their breakdown for a11y and HTML across models (WorkArena L1).}
\label{fig:grounding_error_detailed}
\end{figure}

\begin{table}[t]
\centering
\small
\caption{Grounding error categories.}
\begin{tabularx}{\columnwidth}{@{}lX@{}}
\hline
\textbf{Type} & \textbf{Description} \\
\hline
\multicolumn{2}{@{}p{\columnwidth}@{}}{\textbf{API Format Errors}} \\
Scroll & Invalid argument passed to \texttt{scroll()} \\
Click & Invalid argument passed to \texttt{click()} \\
Select & Invalid argument passed to \texttt{select\_option()} \\
\hline
\multicolumn{2}{@{}p{\columnwidth}@{}}{\textbf{Element Availability Errors}} \\
Not Found & The specified element does not exist in the DOM \\
Not Visible & The specified element is hidden via CSS \\
Not Enabled & The specified element is disabled (e.g., via the \texttt{disabled} attribute) \\
Wrong Type & The specified element does not accept the requested operation \\
Option Not Found & The option value specified in \texttt{select\_option()} does not exist in the element \\
\hline
\multicolumn{2}{@{}p{\columnwidth}@{}}{\textbf{Execution Errors}} \\
Intercepted & A click operation is intercepted by an overlapping element \\
Invalid URL & The URL specified in \texttt{goto()} is invalid \\
\hline
\end{tabularx}
\label{tab:grounding-errors}
\end{table}

\subsection{Grounding Error Analysis}
\label{sec:experiment/grounding_error}

\noindent
\textbf{Setup.}
To further analyze the results of \S\ref{sec:experiment/modality}, 
we examine grounding errors---cases in which an action is not successfully 
accepted by the environment or fails a precondition check prior to execution.
Specifically, we count actions that raise an exception during Playwright execution 
as grounding errors, and classify them based on string patterns in the error messages,
as described in Table~\ref{tab:grounding-errors}.
Errors caused by environment-side issues such as network failures or 
timeouts---where the action itself was correctly formed---are excluded from the analysis.
We compare representative models from each capability group: 
\texttt{gemini-2.5-flash(budget=128/16384)} and \texttt{gpt-5.1(high/low)}, 
for which switching from a11y to HTML improved success rate, 
against \texttt{gpt-oss-20b (high)}, 
which exhibited the largest performance degradation under HTML.

\noindent
\textbf{Error Breakdown.}
Figure~\ref{fig:grounding_error_detailed} shows the number of grounding errors 
and their breakdown for a11y and HTML across models.
When switching from a11y to HTML, the total number of errors decreases for 
the higher-capability models \texttt{gemini-2.5-flash} and \texttt{gpt-5.1}, 
while it increases for the lower-capability model \texttt{gpt-oss-20b (high)}.
This suggests that switching to HTML affects grounding success, which in turn 
contributes to the difference in task success rates. We next examine the breakdown of error types to understand the sources of this difference.
\begin{itemize}
    \item \textbf{Not-found errors} occur when the agent specifies the id of an element that does not exist in the DOM. This type of error increases substantially for \texttt{gpt-oss-20b (high)}, while remaining nearly unchanged for \texttt{gpt-5.1}. Among Gemini models, \texttt{gemini-2.5-flash (budget=128)} shows a large increase, whereas \texttt{gemini-2.5-flash (budget=16384)} successfully suppresses it. These results suggest that either a more capable model or a larger thinking budget helps suppress hallucinated id references under longer inputs.
    \item \textbf{Intercepted errors} occur when a click operation is intercepted by an overlapping element. This type of error decreases across all models, particularly for higher-capability ones. The HTML used in our experiments includes CSS style information specifying z-index, which determines the stacking order of elements. We hypothesize that higher-capability models leverage this information to infer element overlap and thereby avoid such errors.
\end{itemize}

\noindent
\textbf{Summary.}
These findings indicate that higher-capability models can reduce grounding errors by exploiting the rich layout information in HTML, whereas lower-capability models experience increased errors due to hallucination under longer inputs.

\begin{table}[t]
\centering
\small
\tabcolsep 3pt
\caption{Task success rate (\%) by observation history setting and model, evaluated on Workarena L1. Numbers in parentheses indicate the difference from hist0. hist0: no history; hist4/hist9: history from the past 4/9 steps. In the hist9 column, full denotes adding the complete observation history, and diff denotes adding only the diff from the previous step. Values in parentheses after model names indicate \texttt{reasoning\_effort} (high/low) or \texttt{thinking\_budget}. Input token counts are provided as reference values for \texttt{gpt-5.1 (high)}.}
\label{tab:history_workarena_l1_compact}
\begin{tabular}{l|rrr}
\hline
 & hist0 & hist4 & hist9 full/diff \\
n\_input\_tokens & 6720 & 26184 & 39011 / 13670 \\
\hline
\multicolumn{4}{l}{\textbf{Higher-Capability Models (Proprietary)}} \\
\hline
gpt-5.1-2025-11-13 (high) & 55.8 & 58.8 (\textcolor{blue}{+3.0}) & 58.8 (\textcolor{blue}{+3.0}) / 58.8 (\textcolor{blue}{+3.0}) \\
gpt-5.1-2025-11-13 (low) & 49.1 & 53.0 (\textcolor{blue}{+3.9}) & 50.9 (\textcolor{blue}{+1.8}) / 53.3 (\textcolor{blue}{+4.2}) \\
o3-mini-2025-01-31 (high) & 39.7 & 43.0 (\textcolor{blue}{+3.3}) & 43.3 (\textcolor{blue}{+3.6}) / 46.1 (\textcolor{blue}{+6.4}) \\
gemini-2.5-flash (budget=16384) & 45.5 & 48.2 (\textcolor{blue}{+2.7}) & 50.0 (\textcolor{blue}{+4.5}) / 48.2 (\textcolor{blue}{+2.7}) \\
gemini-2.5-flash (budget=128) & 28.5 & 39.4 (\textcolor{blue}{+10.9}) & 39.4 (\textcolor{blue}{+10.9}) / 33.3 (\textcolor{blue}{+4.8}) \\
\hline
\multicolumn{4}{l}{\textbf{Lower-Capability Models (Open-Source)}} \\
\hline
gpt-oss-120b (high) & 46.7 & 49.1 (\textcolor{blue}{+2.4}) & 48.5 (\textcolor{blue}{+1.8}) / 46.4 (\textcolor{red}{-0.3}) \\
gpt-oss-120b (low) & 33.6 & 37.6 (\textcolor{blue}{+4.0}) & 40.0 (\textcolor{blue}{+6.4}) / 39.1 (\textcolor{blue}{+5.5}) \\
gpt-oss-20b (high) & 46.4 & 46.7 (\textcolor{blue}{+0.3}) & 48.8 (\textcolor{blue}{+2.4}) / 49.1 (\textcolor{blue}{+2.7}) \\
gpt-oss-20b (low) & 20.0 & 23.9 (\textcolor{blue}{+3.9}) & 23.9 (\textcolor{blue}{+3.9}) / 24.2 (\textcolor{blue}{+4.2}) \\
\hline
\end{tabular}
\end{table}

\subsection{Observation History and Task Success Rate}
\label{sec:experiment/history}
We investigate the effect of incorporating past observation history.
The observation representation is fixed to a11y, and we compare three settings: no history (hist0), history from the past 4 steps (hist4), and history from the past 9 steps (hist9).
For hist9, we further compare two variants: adding the full observation (full) versus adding only the character-level diff from the previous observation (diff).
Results are shown in Table~\ref{tab:history_workarena_l1_compact}.

\noindent
\textbf{Effect of Observation History.}
The results show that adding observation history improves performance for most models and settings.
Furthermore, \texttt{o3-mini} and \texttt{gemini-2.5-flash} show additional gains from hist9 over hist4, suggesting further room for improvement with longer histories.

\noindent
\textbf{Effect of Diff-Based Observation History.}
Comparing the full and diff variants of hist9, we find that the diff format also generally improves over hist0.
Moreover, for \texttt{gpt-5.1 (low)} and \texttt{o3-mini}, the diff format achieves performance comparable to or better than the full format.
Since the diff format reduces input token count to approximately one-third, it constitutes an efficient alternative to full history.

\begin{figure}[t]
\centering
\includegraphics[width=0.90\columnwidth]{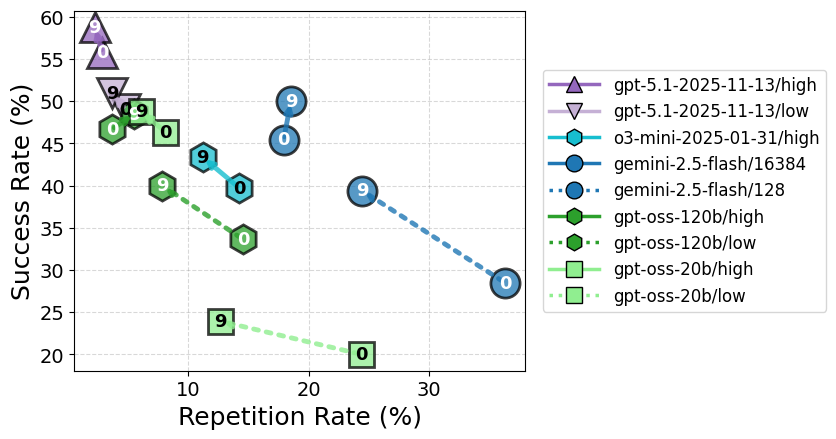}
\caption{Relationship between action repetition rate and task success rate (WorkArena L1). The x-axis shows the rate of actions identical to the previous step; the y-axis shows the task success rate. The number on each data point indicates the length of observation history included as input.}
\label{fig:analysis_repetition}
\end{figure}

\noindent
\textbf{Analysis of Action Repetition}
To investigate why observation history improves performance, we measure the rate at which the agent repeats the same action as in the previous step (repetition rate).
As shown in Figure~\ref{fig:analysis_repetition}, a lower repetition rate is associated with a higher success rate, and adding observation history tends to reduce the repetition rate, particularly when the amount of thinking tokens is small.
This suggests that observation history helps the agent better track task progress and avoid redundant actions.

\section{Related Work}
\label{sec:expanded_related_work}
\noindent
\textbf{Observation Representation and Reduction. }
In web agents, web pages are commonly represented as HTML, which tends to be 
extremely long, posing challenges in terms of LLM context length constraints and 
inference efficiency.
Consequently, existing work has widely adopted observation reduction and selection strategies.
\citet{related_work/mind2web} proposed a framework that trains a cross-encoder 
to score the relevance between the task instruction and each HTML element, 
feeding only the top-$k$ elements to the LLM.
Similarly, \citet{related_work/html-t5} adopt a two-stage pipeline in which a 
dedicated model first extracts and summarizes relevant elements from HTML 
before passing them to the LLM.
\citet{related_work/weblinx} proposed a similar ranking approach but using a 
dual encoder to score the relevance between DOM elements and the full action history.
\citet{related_work/prune4web} proposed generating a Python scoring program from the current sub-task to filter HTML elements programmatically, avoiding direct LLM processing of the full HTML.
Besides HTML-based approaches, several studies have adopted the accessibility tree (a11y), a more compact alternative to HTML.
\citet{related_work/agent_occam} showed that simplifying the a11y by removing task-irrelevant elements improves agent performance.
\citet{related_work/lcow} proposed contextualizing a11y observations by describing the role of each element in natural language.
\citet{related_work/focusagent} proposed using a smaller LLM to selectively extract task-relevant lines from a11y observations.
Research using screenshots only~\citep{related_work/autogui, related_work/seeclick, related_work/clickagent} and work combining screenshots with textual observations~\citep{related_work/webvoyager, related_work/agent_s, related_work/omniparser, related_work/weblinx} have also been proposed; in all cases, compressed or filtered representations are used for the textual observations.

\noindent
\textbf{Observation History and Reduction. }
Leveraging past observations and actions as history in addition to the current observation has been shown to improve performance~\citep{related_work/chain_of_memory}.
However, retaining the full history is impractical, and various strategies have been explored to control the amount of history information.
\citet{related_work/autowebglm} reduce redundancy by storing only action logs and local HTML snippets as history.
\citet{related_work/agent_occam} retain only observations from steps relevant to the currently active subtask.

\noindent
\textbf{Summary.}
In existing work, reducing the amount of information has been the dominant design choice for both single-step observations and observation history.
However, whether such reduction remains beneficial for models with strong long-context capabilities or with larger thinking token budgets has not been sufficiently examined.

\section{Conclusion}
\label{sec:conclusion}
We revisited the trend of observation reduction in web agents and demonstrate that \textbf{observation reduction is not always beneficial for task performance---the optimal observation representation depends on model capability and thinking token budget}: higher-capability models with larger thinking budgets benefit from detailed HTML observations, with gains of up to 17.5 percentage points over compact accessibility tree (a11y) representations, while the benefit grows further as the thinking token budget increases. Our error analysis suggests that stronger models are able to exploit layout-related cues available in HTML---such as CSS z-index information---to avoid grounding errors, while weaker models tend to hallucinate element references under longer inputs. Unlike observation representation, incorporating observation history consistently improves performance across almost all models and settings, and a diff-based history representation offers a token-efficient alternative.

These findings motivate the following suggestions for web agent system design.
\begin{itemize}
    \item \textbf{Adaptively select observation representation.} We recommend selecting the observation representation based on model capability and reasoning budget. For high-capability models with larger thinking budgets, the performance gains from richer HTML observations may well outweigh the increased inference cost. Furthermore, as shown in \S\ref{sec:experiment/modality}, the benefit of HTML varies by task category, suggesting that task characteristics should also be taken into account.
    \item \textbf{Incorporate observation history.} We encourage incorporating observation history; when inference cost is a concern, diff-based history offers a practical token-efficient alternative.
\end{itemize}

\section*{Limitation}
\label{sec:limitation}
\begin{itemize}
    \item \textbf{Generalizability across websites and domains.} Our experiments are conducted on WorkArena L1, a benchmark built on a real-world website (ServiceNow), which provides a degree of practical validity. Nevertheless, whether the findings generalize to other websites and domains remains unverified.
    \item \textbf{Grounding scheme.} Our experiments are limited to id-based grounding; whether similar trends hold under coordinate-based or other grounding schemes is an open question.
    \item \textbf{Mechanism of HTML benefit.} Our error analysis suggests that CSS layout cues contribute to the HTML benefit, but this has not been directly verified through ablation and additional factors may be involved.
    \item \textbf{Observation history with richer representations.} Our history experiments fix the observation representation to a11y, leaving the effect of combining richer observations such as HTML with history unexplored. Furthermore, WorkArena L1 involves tasks of up to 15 steps, and the effect of observation history on longer-horizon tasks remains unknown.
\end{itemize}

\bibliography{main}

@inproceedings{react,
  author       = {Shunyu Yao and others},
  title        = {ReAct: Synergizing Reasoning and Acting in Language Models},
  booktitle    = {The Eleventh International Conference on Learning Representations,
                  {ICLR} 2023, Kigali, Rwanda, May 1-5, 2023},
  year         = {2023},
  url          = {https://openreview.net/forum?id=WE\_vluYUL-X},
}

@article{related_work/chain_of_memory,
  author       = {Xinzge Gao and others},
  title        = {Chain-of-Memory: Enhancing {GUI} Agents for Cross-Application Navigation},
  journal      = {CoRR},
  volume       = {abs/2506.18158},
  year         = {2025},
  url          = {https://doi.org/10.48550/arXiv.2506.18158},
  doi          = {10.48550/ARXIV.2506.18158},
  eprinttype   = {arXiv},
  eprint       = {2506.18158},
  bibsource    = {dblp computer science bibliography, https://dblp.org}
}

@article{related_work/focusagent,
  author       = {Imene Kerboua and others},
  title        = {FocusAgent: Simple Yet Effective Ways of Trimming the Large Context
                  of Web Agents},
  journal      = {CoRR},
  volume       = {abs/2510.03204},
  year         = {2025},
  url          = {https://doi.org/10.48550/arXiv.2510.03204},
  doi          = {10.48550/ARXIV.2510.03204},
  eprinttype   = {arXiv},
  eprint       = {2510.03204},
  bibsource    = {dblp computer science bibliography, https://dblp.org}
}

@inproceedings{related_work/prune4web,
  author       = {Jiayuan Zhang and others},
  title        = {Prune4Web: {DOM} Tree Pruning Programming for Web Agent},
  booktitle    = {Fortieth {AAAI} Conference on Artificial Intelligence, Thirty-Eighth
                  Conference on Innovative Applications of Artificial Intelligence,
                  Sixteenth Symposium on Educational Advances in Artificial Intelligence,
                  {AAAI} 2026, Singapore, January 20-27, 2026},
  pages        = {34710--34718},
  publisher    = {{AAAI} Press},
  year         = {2026},
  url          = {https://doi.org/10.1609/aaai.v40i41.40772},
}

@inproceedings{related_work/oscar,
  author       = {Xiaoqiang Wang and
                  Bang Liu},
  title        = {{OSCAR:} Operating System Control via State-Aware Reasoning and Re-Planning},
  booktitle    = {The Thirteenth International Conference on Learning Representations,
                  {ICLR} 2025, Singapore, April 24-28, 2025},
  year         = {2025},
  url          = {https://openreview.net/forum?id=VuTrZzrPfn},
}

@inproceedings{related_work/autowebglm,
  author       = {Hanyu Lai and others},
  title        = {AutoWebGLM: {A} Large Language Model-based Web Navigating Agent},
  booktitle    = {Proceedings of the 30th {ACM} {SIGKDD} Conference on Knowledge Discovery
                  and Data Mining, {KDD} 2024, Barcelona, Spain, August 25-29, 2024},
  pages        = {5295--5306},
  year         = {2024},
  url          = {https://doi.org/10.1145/3637528.3671620},
  doi          = {10.1145/3637528.3671620},
}

@article{related_work/omniparser,
  author       = {Yadong Lu and others},
  title        = {OmniParser for Pure Vision Based {GUI} Agent},
  journal      = {CoRR},
  volume       = {abs/2408.00203},
  year         = {2024},
  url          = {https://doi.org/10.48550/arXiv.2408.00203},
  doi          = {10.48550/ARXIV.2408.00203},
  eprinttype    = {arXiv},
  eprint       = {2408.00203},
}

@inproceedings{related_work/agent_s,
  author       = {Saaket Agashe and others},
  title        = {Agent {S:} An Open Agentic Framework that Uses Computers Like a Human},
  booktitle    = {The Thirteenth International Conference on Learning Representations,
                  {ICLR} 2025, Singapore, April 24-28, 2025},
  year         = {2025},
  url          = {https://openreview.net/forum?id=lIVRgt4nLv},
}

@inproceedings{related_work/webvoyager,
    title = "{W}eb{V}oyager: Building an End-to-End Web Agent with Large Multimodal Models",
    author = "He, Hongliang and others",
    booktitle = "Proceedings of the 62nd Annual Meeting of the Association for Computational Linguistics (Volume 1: Long Papers)",
    month = aug,
    year = "2024",
    url = "https://aclanthology.org/2024.acl-long.371/",
    doi = "10.18653/v1/2024.acl-long.371",
    pages = "6864--6890",
}

@inproceedings{related_work/clickagent,
    title = "{C}lick{A}gent: Enhancing {UI} Location Capabilities of Autonomous Agents",
    author = "Hoscilowicz, Jakub  and
      Janicki, Artur",
    booktitle = "Proceedings of the 26th Annual Meeting of the Special Interest Group on Discourse and Dialogue",
    month = aug,
    year = "2025",
    url = "https://aclanthology.org/2025.sigdial-1.38/",
    pages = "471--476",
}

@inproceedings{related_work/seeclick,
  author       = {Kanzhi Cheng and others},
  title        = {SeeClick: Harnessing {GUI} Grounding for Advanced Visual {GUI} Agents},
  booktitle    = {Proceedings of the 62nd Annual Meeting of the Association for Computational
                  Linguistics (Volume 1: Long Papers), {ACL} 2024, Bangkok, Thailand,
                  August 11-16, 2024},
  pages        = {9313--9332},
  year         = {2024},
  url          = {https://doi.org/10.18653/v1/2024.acl-long.505},
  doi          = {10.18653/V1/2024.ACL-LONG.505},
}

@inproceedings{related_work/autogui,
  author       = {Hongxin Li and others},
  title        = {AutoGUI: Scaling {GUI} Grounding with Automatic Functionality Annotations
                  from LLMs},
  booktitle    = {Proceedings of the 63rd Annual Meeting of the Association for Computational
                  Linguistics (Volume 1: Long Papers), {ACL} 2025, Vienna, Austria,
                  July 27 - August 1, 2025},
  pages        = {10323--10358},
  year         = {2025},
  url          = {https://aclanthology.org/2025.acl-long.510/},
}

@inproceedings{related_work/webagent_survey,
author = {Ning, Liangbo and others},
title = {A Survey of WebAgents: Towards Next-Generation AI Agents for Web Automation with Large Foundation Models},
year = {2025},
url = {https://doi.org/10.1145/3711896.3736555},
doi = {10.1145/3711896.3736555},
booktitle = {Proceedings of the 31st ACM SIGKDD Conference on Knowledge Discovery and Data Mining V.2},
pages = {6140-6150},
}

@article{related_work/LCLM_reasoning/beyond_isolated,
  author       = {Yifei Wang},
  title        = {Beyond Isolated Capabilities: Bridging Long CoT Reasoning and Long-Context
                  Understanding},
  journal      = {CoRR},
  volume       = {abs/2507.14849},
  year         = {2025},
  url          = {https://doi.org/10.48550/arXiv.2507.14849},
  doi          = {10.48550/ARXIV.2507.14849},
  eprinttype    = {arXiv},
  eprint       = {2507.14849},
}

@article{related_work/LCLM_reasoning/cot_matters,
  author       = {Dawei Zhu and others},
  title        = {Chain-of-Thought Matters: Improving Long-Context Language Models with
                  Reasoning Path Supervision},
  journal      = {CoRR},
  volume       = {abs/2502.20790},
  year         = {2025},
  url          = {https://doi.org/10.48550/arXiv.2502.20790},
  doi          = {10.48550/ARXIV.2502.20790},
  eprinttype    = {arXiv},
  eprint       = {2502.20790},
}

@inproceedings{related_work/mind2web,
  author       = {Xiang Deng and others},
  title        = {Mind2Web: Towards a Generalist Agent for the Web},
  booktitle    = {Advances in Neural Information Processing Systems 36: Annual Conference
                  on Neural Information Processing Systems 2023, NeurIPS 2023, New Orleans,
                  LA, USA, December 10 - 16, 2023},
  year         = {2023},
  url          = {http://papers.nips.cc/paper\_files/paper/2023/hash/5950bf290a1570ea401bf98882128160-Abstract-Datasets\_and\_Benchmarks.html},
}

@inproceedings{related_work/html-t5,
  author       = {Izzeddin Gur and others},
  title        = {A Real-World WebAgent with Planning, Long Context Understanding, and
                  Program Synthesis},
  booktitle    = {The Twelfth International Conference on Learning Representations,
                  {ICLR} 2024, Vienna, Austria, May 7-11, 2024},
  year         = {2024},
  url          = {https://openreview.net/forum?id=9JQtrumvg8},
}

@inproceedings{related_work/weblinx,
  author       = {Xing Han L{\`{u}} and
                  Zdenek Kasner and
                  Siva Reddy},
  title        = {WebLINX: Real-World Website Navigation with Multi-Turn Dialogue},
  booktitle    = {Forty-first International Conference on Machine Learning, {ICML} 2024,
                  Vienna, Austria, July 21-27, 2024},
  year         = {2024},
  url          = {https://openreview.net/forum?id=mUSPhG4uDW},
}

@inproceedings{related_work/agent_occam,
  author       = {Ke Yang and others},
  title        = {AgentOccam: {A} Simple Yet Strong Baseline for LLM-Based Web Agents},
  booktitle    = {The Thirteenth International Conference on Learning Representations,
                  {ICLR} 2025, Singapore, April 24-28, 2025},
  year         = {2025},
  url          = {https://openreview.net/forum?id=oWdzUpOlkX},
}

@inproceedings{related_work/lcow,
  author       = {Dongjun Lee and others},
  title        = {Learning to Contextualize Web Pages for Enhanced Decision Making by
                  {LLM} Agents},
  booktitle    = {The Thirteenth International Conference on Learning Representations,
                  {ICLR} 2025, Singapore, April 24-28, 2025},
  year         = {2025},
  url          = {https://openreview.net/forum?id=3Gzz7ZQLiz},
}

@article{browsergym,
    title={The BrowserGym Ecosystem for Web Agent Research},
    author={Thibault Le Sellier de Chezelles and others},
    journal={Transactions on Machine Learning Research},
    issn={2835-8856},
    year={2025},
    url={https://openreview.net/forum?id=5298fKGmv3},
}

@inproceedings{benchmark/workarena,
  author       = {Alexandre Drouin and others},
  title        = {WorkArena: How Capable are Web Agents at Solving Common Knowledge
                  Work Tasks?},
  booktitle    = {Forty-first International Conference on Machine Learning, {ICML} 2024,
                  Vienna, Austria, July 21-27, 2024},
  year         = {2024},
  url          = {https://openreview.net/forum?id=BRfqYrikdo},
}

@inproceedings{benchmark/workarena++,
  author       = {L{\'{e}}o Boisvert and others},
  title        = {WorkArena++: Towards Compositional Planning and Reasoning-based Common
                  Knowledge Work Tasks},
  booktitle    = {Advances in Neural Information Processing Systems 38: Annual Conference
                  on Neural Information Processing Systems 2024, NeurIPS 2024, Vancouver,
                  BC, Canada, December 10 - 15, 2024},
  year         = {2024},
  url          = {http://papers.nips.cc/paper\_files/paper/2024/hash/0b82662b6c32e887bb252a74d8cb2d5e-Abstract-Datasets\_and\_Benchmarks\_Track.html},
}

@article{models/gemini_25,
  author       = {Gemini Team},
  title        = {Gemini 2.5: Pushing the Frontier with Advanced Reasoning, Multimodality,
                  Long Context, and Next Generation Agentic Capabilities},
  journal      = {CoRR},
  volume       = {abs/2507.06261},
  year         = {2025},
  url          = {https://doi.org/10.48550/arXiv.2507.06261},
  doi          = {10.48550/ARXIV.2507.06261},
  eprinttype    = {arXiv},
  eprint       = {2507.06261},
}

@article{models/gpt-oss,
  author       = {OpenAI},
  title        = {gpt-oss-120b {\&} gpt-oss-20b Model Card},
  journal      = {CoRR},
  volume       = {abs/2508.10925},
  year         = {2025},
  url          = {https://doi.org/10.48550/arXiv.2508.10925},
  doi          = {10.48550/ARXIV.2508.10925},
  eprinttype    = {arXiv},
  eprint       = {2508.10925},
}

@article{models/llama3,
  author       = {Llama Team},
  title        = {The Llama 3 Herd of Models},
  journal      = {CoRR},
  volume       = {abs/2407.21783},
  year         = {2024},
  url          = {https://doi.org/10.48550/arXiv.2407.21783},
  doi          = {10.48550/ARXIV.2407.21783},
  eprinttype    = {arXiv},
  eprint       = {2407.21783},
}

@article{models/qwen3,
  author       = {An Yang and others},
  title        = {Qwen3 Technical Report},
  journal      = {CoRR},
  volume       = {abs/2505.09388},
  year         = {2025},
  url          = {https://doi.org/10.48550/arXiv.2505.09388},
  doi          = {10.48550/ARXIV.2505.09388},
  eprinttype    = {arXiv},
  eprint       = {2505.09388},
}
\bibliographystyle{colm2026_conference}

\end{document}